# Root Cause Analysis Of Productivity Losses In Manufacturing Systems Utilizing Ensemble Machine Learning

Jonas Gram[1], Brandon K. Sai[1], Thomas Bauernhansl[1,2]
[1]*Fraunhofer Institute for Manufacturing Engineering and Automation IPA, Stuttgart, Germany*
[2]*Institute of Industrial Manufacturing and Management, IFF, Stuttgart, Germany*

## Abstract

In today's rapidly evolving landscape of automation and manufacturing systems, the efficient resolution of productivity losses is paramount. This study introduces a data-driven ensemble approach, utilizing the cyclic multivariate time series data from binary sensors and signals from Programmable Logic Controllers (PLCs) within these systems. The objective is to automatically analyze productivity losses per cycle and pinpoint their root causes by assigning the loss to a system element.

The ensemble approach introduced in this publication integrates various methods, including information theory and machine learning behavior models, to provide a robust analysis for each production cycle.

To expedite the resolution of productivity losses and ensure short response times, stream processing becomes a necessity. Addressing this, the approach is implemented as data-stream analysis and can be transferred to batch processing, seamlessly integrating into existing systems without the need for extensive historical data analysis. This method has two positive effects. Firstly, the result of the analysis ensures that the period of lower productivity is reduced by identifying the likely root cause of the productivity loss. Secondly, these results are more reliable due to the ensemble approach and therefore avoid dependency on technical experts.

The approach is validated using a semi-automated welding manufacturing system, an injection molding automation system, and a synthetically generated test PLC dataset. The results demonstrate the method's efficacy in offering a data-driven understanding of process behavior and mark an advancement in autonomous manufacturing system analysis.

## 1. Introduction

**Context and Motivation.** Within the *Digital Transformation* in manufacturing, the integration of *cyber-physical systems (CPS)* stands out as a pivotal development. CPS, characterized by the convergence of digital and physical elements, are augmented by digital services that fulfill the 'cyber' aspect of the concept. These services cover particular areas such as maintenance or optimization, with growing relevance for operations [1]. The optimization of automated production systems faces an additional challenge as production processes become increasingly dynamic. For instance, while order-related potential for improvement may be identified, it might remain unused since the production order with a small lot size has already been completed. This dynamic results in narrower windows of opportunity for intervention and emphasizes the urgency of an adequate response time. Furthermore, the criticality of response time is evident, as highlighted by metrics such as *Mean Time To Repair (MTTR)* [2]. The most time-consuming task within optimization is diagnostic analysis. Due to its reliance on expert knowledge and contextual understanding, the diagnostic analysis is predominantly conducted manually. This dependence on manual tasks shows unexploited potential that will allow an automated system to evolve into a CPS. The aim of diagnostic analysis is to determine which component of the system causes the detected loss of productivity. With this, losses can be allocated at component level, and conversely, countermeasures can be taken for optimization.

This paper aligns with the overarching research work on identifying productivity losses in automated production systems using behavioral models [3,4]. This publication specifically focuses on the diagnostic analysis performed by an automated, unsupervised ensemble method, adapting to the dynamic nature of modern manufacturing systems. By employing an ensemble approach, root cause identification becomes more reliable since the approach seeks to cover a wide spectrum of possible solution paths. This is in line with the overall problem definition, as the optimization itself is less idealistic and more pragmatic [5].

**Contributions.**

An ensemble framework for analyzing cyclic multivariate time series data from manufacturing systems sensors and PLCs is proposed, automatically pinpointing root cause features for cycles with high productivity loss in manufacturing settings.

The ensemble's architecture is engineered to ensure robustness and reduce false-positive outputs, leveraging behavioral modeling, as well as incremental and continual learning.

The framework allows for parallelization of its components and offers a customizable balance between response times and depth of contextual analysis. The method is validated across diverse manufacturing and automation system settings, demonstrating the potential for broad application in the manufacturing industry.

## 2. Related Work

### 2.1 Explaining causality of productivity losses in manufacturing systems

In the exploration of the causality behind productivity losses in manufacturing systems, this work has drawn from foundational works in operational performance assessment. The field of research is well developed as depicted in the work of Muchiri et al. [6]. However, recent activities have been sparked by the digital transformation. Notably, the fundamental work of Slack et al. [2] and the contributions of Tangen [7] and Ungern-Sternberg et al. [8] offer comprehensive insights into the assessment and categorization of operational performance. Slack et al. provide a holistic view of performance assessment, elucidating operational objectives derived from the strategic and societal considerations of an organization. These objectives, encompassing quality, speed, dependability, flexibility, and cost, serve as fundamental benchmarks for evaluating operational effectiveness. A challenge in application is that these objectives do not share the same measurement units. This is substantiated by Tangen's contribution, which introduces three dimensions of measurement for assessing operational performance: output-based, time-based, and monetary-based measures. In combination, five objectives are defined, namely: performance, considering factors such as output quantity, production time, and associated costs. Ungern-Sternberg et al. provide a framework for performance measurement that integrates both, the five objectives and the three dimensions. The result of this framework is a unified performance assessment based on time-series. Central to Ungern-Sternberg's framework is the classification of operational states, facilitating the identification and categorization of productivity losses. By partitioning total time into distinct categories such as order processing time, downtime, idle time, and non-disposable time, the framework provides a structured approach to assessing system performance and identifying areas for improvement. Furthermore, the classification scheme offers insights into the nature of productivity losses, distinguishing between technical and organizational factors contributing to inefficiencies. Through this framework, the distinction between productivity and performance becomes visible, as efficiency can only be evaluated by fulfilling the third requirement of the framework, namely: versatile data integration. Nevertheless, this framework does not specify which data sources are intended. Thus, the resolution of the accuracy of the findings remains undetermined. The highest resolution of data can be found at the PLC level of the automation pyramid. This paper examines whether and to what extent data from the PLC level offers an added value for this framework.

### 2.2 Machine Learning in root cause analysis for anomalies in multivariate time series data

Machine- and Deep Learning techniques have gained increasing importance in identifying and explaining anomalies in diverse time series applications. This publication focuses on leveraging these techniques for root cause analysis of productivity losses in manufacturing environments. To achieve this, specifically data from PLC and sensor systems is analyzed, both being quintessential examples of time series data. As outlined previously, the research gap can be found in the application. Consequently, comparable data characteristics must be included in the methodological research. The structure of these PLC data streams is analogous to those found in other domains, such as cyber security or health care. Hence, an overview about significant research and development in time series analysis for different applications should be given.

With the research conducted by Chen et al., high-order dynamic behaviors are separated from static process characteristics. By implementing a dynamic fault isolation strategy for each dynamic node, their model offers an understanding of the root causes for targeted interventions [9]. Perepu et al. introduced an unsupervised method for root cause analysis of anomalies in dynamic manufacturing systems, utilizing sparse optimization techniques. This approach generates insights without the need for extensive labeled data, which is often a bottleneck in machine learning applications [10]. To further address interpretability, leveraging causal data structure, root cause analysis based on a causal graph being inferred from the data was proposed by Budhathoki et al. The root cause contributions were quantified using information-theoretic outlier scores and Shapley values [11]. In 2023, Assaad et al. present a method for root causes identification of anomalies, utilizing an acyclic summary cause graph representing causal relations in a dynamic system. Based on dividing the root cause identification task into multiple sub-problems, the graph is used to directly identify root causes. Their research focuses specifically on root cause analysis with a focus on interpretability [12]. Modeling available manual information in manufacturing through a knowledge graph, Wehner et al. combine data driven root cause analysis with expert knowledge. The Causal Bayesian Network employed is improved through iterative feedback loops and available domain expertise, while its search space is pruned by utilization of the knowledge graph [13].

### 2.3 Ensemble methods in data-driven analysis of manufacturing systems

Focusing back only on the application field of manufacturing and automation systems, research and development utilizing ensemble models needs to be outlined. In the manufacturing domain, adopting ensemble models has shown great potential for improving the analysis and optimization of various processes. The following section outlines the key research in ensemble methods that has significantly contributed to advancements. Ensemble models for optimizing machining processes in IoT systems were highlighted by Garrido-Labrador et al. (2020). Due to the ability to extract various different information streams from the IoT system, the great performance of ensemble models for the optimization of machining processes was demonstrated [14]. An ensemble approach built from multiple machine learning models was utilized by Jose et al. to analyze vibration data for machine fault diagnosis. They also built a voting classifier to address the problem of weighing multiple single statements within the ensemble, which would otherwise induce high output variance depending on the weights [15]. Kaupp et al. take a novel approach with an AutoEncoder-ensemble framework for unsupervised variable selection to improve focus in fault diagnosis in manufacturing processes. By identifying key variables, their method enhances the precision of unsupervised fault diagnosis systems without knowledge based pre-selection [16]. Lately, ensemble methods have been combined with active learning contextual bandits in the field of decision-making for manufacturing systems by Zeng et al. Here, the ensemble enhances prediction robustness due to aggregating multiple models. Active learning combined with ensemble methods additionally boosts learning efficiency by combining multiple insights to label the most informative data points, achieving higher accuracy [17]. With ensemble methods, analytical results reach higher levels in reliability, a frequently addressed shortcoming of data-driven applications in manufacturing.

## 3. Method

This work presents a method for analyzing cyclic manufacturing time series, combining linear and non-linear feature importance, incremental learning within dynamic manufacturing systems, and the integration of behavioral and structural modeling for root cause analysis of productivity losses. It utilizes ensemble learning to provide fast and robust feedback for manufacturing systems.

### 3.1 Data and Descriptive Analysis

The foundation of this data-driven analysis of manufacturing systems is based on the aforementioned binary PLC and sensor data. Following the principle of automation, a recurring pattern is inherent in the data record, and the duration of this repetition is dependent to a production order [18]. The data encapsulates the operational dynamics of a manufacturing system, with each cycle consisting of a complete set of operations needed for a manufacturing step. Mathematically, the cyclic multivariate time series PLC data is defined as $X^c = (X_1^c, \dots, X_n^c) \in \mathbb{R}^{n \times d}$ with $n$ time steps, dimensionality $d$ and cycle number $c$ all being part of the natural numbers $\mathbb{N}$. The cyclic behavior varies in duration of cycles and operations, but not in the order of operations within each cycle. Table 1 in the appendix demonstrates the structure of the data visually.

The focus of this work lies in the root cause analysis of such cycles, with a measured productivity loss exceeding the norm, to detect potential responsible signals. The outlined approach relies on a manufacturing productivity flag $P = \{0,1\}$ per cycle to be analyzed, but it does not necessarily have to be the exact same measurement in every system. $P = 0$ indicates a low productivity loss cycle, $P = 1$ a high-loss cycle. The flag can be assigned based on descriptive analysis, expert knowledge or data-driven via time series analysis.

### 3.2 Automatic identification of root causes

To thoroughly analyze the cycles with high productivity loss and pinpoint potential root causes, an ensemble approach is developed. Since ensemble methods aggregate the statements of various models, the effect of noise or outliers on the final output is reduced, enhancing its robustness [19]. This is also important to reduce false positive statements due to single model sensitivity, potentially causing downtime in manufacturing. Figure 1 displays the ensemble approach developed.

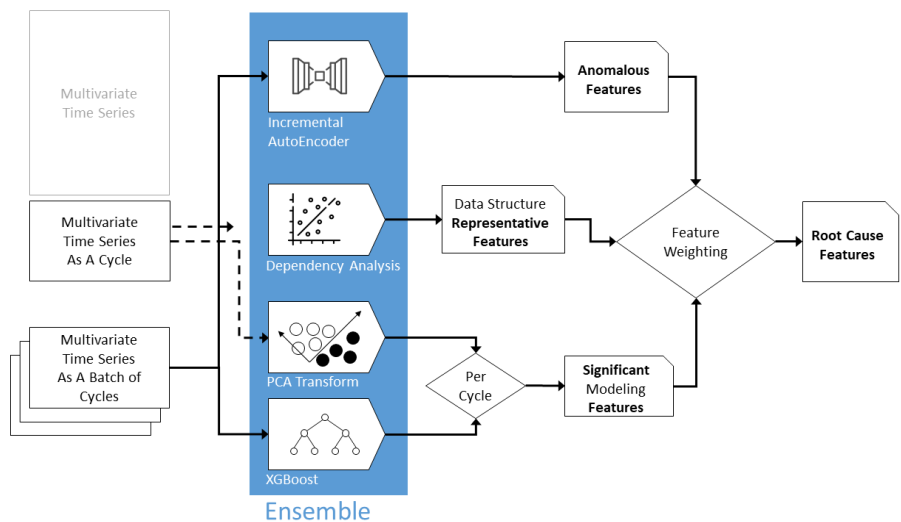

Figure 1: Overview of the ensemble model used to determine root cause features in high productivity loss cycles

The **AutoEncoder** (AE) with **Incremental Learning** (IAE) is a variant of AutoEncoders adapted for data-streams and environments with a dynamic data distribution over time. Analogous to conventional AE, the IAE is designed to learn a non-linear function to encode and subsequently reconstruct input data. During inference, the reconstruction error serves as an anomaly score, indicating manifestations of deviations from the learned and expected normal data structure. The AutoEncoder is defined by [20] as:

$$z = E(X; \theta_e) = \sigma(W_e X + b_e) \quad (1)$$

Where $z$ represents the latent encoding of the input $X$ by the encoder function $E$, parametrized by $\theta_e = \{W_e, b_e\}$ with the weight matrix $W_e$ and the bias vector $b_e$. The activation function is denoted by $\sigma$.

$$\hat{X} = D(z; \theta_d) = \sigma(W_d h + b_d) \quad (2)$$

Where $\hat{X}$ represents the reconstructed input obtained from the latent representation $z$ by the decoder function $D$, parametrized by $\theta_d = \{W_d, b_d\}$ with the weight matrix $W_d$ and the bias vector $b_d$. The activation function is denoted by $\sigma$.

To still address fast response times and granular analysis but enable the AutoEncoder to learn temporal dependencies between manufacturing processes, $a$ cycles will be concatenated into a batch $B_j = [X^j, X^{j+a-1}] \in \mathbb{R}^{(an) \times d}$ starting from a specific cycle $c = j$ where $j$ is a natural number indicating the starting cycle for the batch. The number of cycles $a$ within each batch depends on the cycle time and customer needs.

The uniqueness of the IAE is the training on segments of cyclic time data, one batch at a time. For each batch, the IAE updates its parameters in a training iteration to incorporate the new information without forgetting and overwriting the previously learned patterns. This incremental training enables learning long-term behavior while still strongly weighing the newest cycles and not being reliant on historical data.

To ensure continual learning over progressing cycles without catastrophic forgetting [21], two different techniques are applied. The *replay buffer* $\mathcal{R}$ stores up to 100 previously analyzed batches. Every $r \in \mathbb{N}$ seen batches, the model undergoes a selective training iteration on $\left\lceil \frac{len(\mathcal{R})}{8} \right\rceil + 3$ batches randomly sampled from $\mathcal{R}$, where $len(\mathcal{R})$ denotes the current number of batches stored in $\mathcal{R}$.

Additionally, Elastic Weight Consolidation (EWC) [22] loss $\mathcal{L}_{EWC}$ is employed to retain previously acquired knowledge when updating the model parameters during training. The EWC approach leverages the Fisher Information matrix $F$, described in [23], which quantifies the importance of each model parameter with respect to the performance on previous batches through the parameter $\lambda$ which balances the trade-off between fitting new data and retaining former knowledge. The Fisher matrix represents a second-order approximation to the model's error surface around the optimal parameters, thus providing a way to penalize significant deviations from these parameters during new learning phases. This helps in preventing catastrophic forgetting by adding a regularization term that constrains the parameter updates, particularly those parameters crucial for previous tasks' performance.

Based on this, the AutoEncoder loss function $\mathcal{L}_{AE}$ calculates the reconstruction loss between the original input $B_j$ and the reconstructed output $\hat{B}_j$ and is combined with $\mathcal{L}_{EWC}$, incorporating the Fisher matrix $F$ and the weighing parameter $\lambda$ to prevent forgetting important information.

$$\mathcal{L} = \mathcal{L}_{AE}(B_j, \hat{B}_j) + \mathcal{L}_{EWC} = \frac{1}{n}\sum_{i=1}^{n}\|B_i - \hat{B}_i\|^2 + \sum_k \frac{\lambda}{2} F_k (\theta_k - \theta_{A,k}^*)^2 \quad (3)$$

$\theta_A^*$ denotes the approximate Gaussian distribution of the model parameters and $k$ represents the labels per parameter. For each learning step, the model tries to minimize $\mathcal{L}$ and updates its parameters accordingly. After each training iteration, and only for cycles with high productivity loss, the IAE reconstructs the latest cycle during inference. The reconstruction error per feature can potentially be linked to productivity losses, while definitely revealing deviations from previously learned behavior [24]. The use of IAE addresses the dynamic nature of manufacturing data, adapting to new patterns or changes in operational behavior. This adaptability is essential for accurately identifying anomalies over time, which are potential indicators of root causes for productivity losses. Based on the most significant anomaly scored features, the interim results $I_{1a}^c$ containing the top $t$ anomaly locations and $I_{1b}^c$ containing the next $t$ features are returned.

Within the dependency analysis, **Pearson's Correlation Coefficient** (PCC) helps the ensemble filter out irrelevant features per cycle $X^c$, as well as capturing linear relationships between features.
**Mutual Information** (MI) quantifies the information gained about one random variable through another. MI helps the ensemble uncover nonlinear pairwise relationships between sensor readings or PLC signals, going beyond the linear correlations that PCC captures. For discrete features per cycle $X_i^c$ and $X_{i+1}^c$ it is defined by Shannon [25] as:

$$I(X_i; X_{i+1}) = \sum_{x_i \in X_i} \sum_{x_{i+1} \in X_{i+1}} p(x_i, x_{i+1}) \log \left( \frac{p(x_i, x_{i+1})}{p(x_i) p(x_{i+1})} \right) \qquad (4)$$

Where $p(x_i, x_{i+1})$ is the joint probability distribution function, $p(x_i)$ and $p(x_{i+1})$ denote the marginal probability distribution functions for $x_i$ and $x_{i+1}$ respectively. The goal is to compute the features with the highest information overlap in the data and the highest amount of dependable other signals. These consist of representative features in the data structure, potentially pinpointing bottlenecks and other important parts in the underlying systems structure. To achieve this, the union of the most significant features from the PCC and MI is taken as the first ensemble step and returned as the interim result of feature indices $I_2^c$.

Lastly, **Principal Component Analysis** (PCA) for structural embedding and **eXtreme Gradient Boosting** (XGBoost) for cycle behavior modeling was utilized. PCA is computed per cycle $X^c$ and reduces the dimensionality of the data by transforming it into a set of linearly uncorrelated principal components that capture maximum variance. This transformation highlights the underlying structure of the data through a low-rank structure matrix, emphasizing the most significant features [26], potentially correlating with productivity losses.

XGBoost is an advanced implementation of gradient boosting, an ensemble technique which uses multiple decision trees sequentially. Each tree in XGBoost is built to minimize a loss function, with focus on correcting the mistakes of the preceding tree. This is achieved by gradient descent on the loss function, allowing the model to incrementally improve and adapt to the complex patterns in the data. The algorithm was originally introduced by [27] and developed further by [28]. The model is trained on the same time series batch $B_j$ as the AutoEncoder, with the addition of each cycle within the batch being labeled based on the productivity loss flag $P = \{0,1\}$. In this way, the model learns to distinguish between these two states and learns normal behavior, not leading to productivity loss. XGBoost is capable of handling complex relationships within high-dimensional data in a supervised way, and thus builds a complement to the simpler approach of correlation analysis introduced previously. The model updates its parameters as shown in equation 5.

$$Obj(\Theta) = \sum_{i=1}^n l(y_i, \hat{y}_i) + \sum_{k=1}^K \Omega(f_k) \qquad (5)$$

With $\Theta$ representing the parameters of the model, $l$ being a differentiable convex loss function measuring the difference between the predicted label $\hat{y}_i$, and the actual label $y_i$ per data point. $\Omega$ being the regularization term, penalizing the model's complexity, $K$ the number of trees, and $f_k$ the individual trees in the model.
To identify a feature set for the root cause analysis, feature importance is computed after each XGBoost modeling per batch of cycles. The feature importance metric is derived from a combination of factors as outlined in the implementation of [28]. Per high loss cycle, the intersection of these XGBoost behavior model important features with the most prominent PCA structural modeling features is computed to return $I_3^c$, consisting of features pivotal for modeling the datasets structure as well as the productivity loss behavior over the cycles. Based on the three subsets of features with a high impact in cycle-based root cause analysis, a mutual root cause output is formed. The initial root cause score per feature per cycle is denoted as $s^c(f) = 0$. The ensemble model output is given by this score based on equation 6.

$$s^c(f) = \sum_{m \in I_{1a}^c} 2 * \delta_{m,f} + \sum_{m \in I_{1b}^c} 1 * \delta_{m,f} + \sum_{m \in I_2^c} 1 * \delta_{m,f} + \sum_{m \in I_3^c} 1 * \delta_{m,f} \quad (6)$$

$\delta_{m,f}$ is the Kronecker delta, equal to 1 if $m = f$ and 0 otherwise. Depending on the index subset a feature belongs to, the score for feature $f$ increases by 2 for $I_{1a}^c$ and by 1 for the others.

A feature $f$ is considered a potential root cause if $s^c(f) \geq 2$, with higher scores indicating a higher importance of the feature $f$ in the analysis.

## 4. Validation and Discussion

### 4.1 Case Study: Semi-automated welding manufacturing system

This system with multiple PLC's and manual workstations has been observed for one week. 20 relevant signals have been recorded, over 2381 cycles with 94 seconds ideal cycle time and each cycle consisting of 13 different states. The extended observation period, paired with the systems inherent process variance and planned downtimes as well as disruptions and anomalies, presents a complex scenario with high variance for analysis. This complexity is indicative of many semi-automated manufacturing systems where human intervention is still highly significant. Figure 2 displays the root cause analysis using the proposed ensemble method, highlighting significant features with scores $\geq 2$ in high-loss cycles of complex manufacturing systems. The graph uses color-coding to identify key features likely cause productivity losses, correctly pinpointing the anonymized features 12 and 15 as the main contributors. These two sensors are associated with critical manual processes, each leading to high waiting times for the automation system.

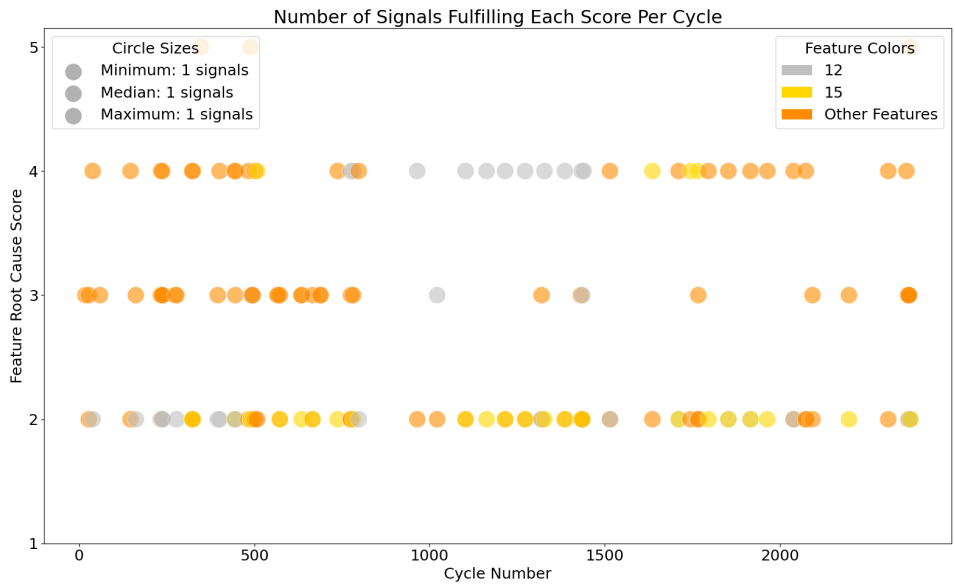

Figure 2: Assigned feature root cause scores over the duration of all cycles: The number of signals holding that score would be represented by larger circles. The color encoding shows the different features being detected as root cause. This shows that the approach can detect potential root causes in high loss cycles.

### 4.2 Case Study: Injection molding automation system

This fully automated injection molding machine has 83 associated binary signals, recorded over 40 minutes. The data log captures 32 cycles with each cycle comprising 18 different states and an expected cycle time of 70 seconds. This process represents highly automated manufacturing environments where precision and speed are paramount and the margin for error as well as process variance are minimal. The limited observation cycles, high signal precision and low downtimes offer a contrasting scenario to the semi-automated system. Figure 3 in the appendix shows the root cause assignment per high loss cycle of this system, identifying key root cause features. This demonstrates the ability of the model to isolate a feature group that most likely causes high productivity loss in high precision automation systems.

### 4.3 Automation system simulation via synthetic PLC data

This automation system was simulated by observing 26 PLC signals over 36 hours. The log consists of 400 cycles, with an ideal cycle time of 140 seconds and 14 different operational steps per cycle. The dataset serves as a controlled environment to test the robustness and versatility of the ensemble machine learning approach. By designing a synthetic dataset, the method is validated against known parameters and conditions, providing a labeled baseline for the effectiveness of the analysis. The operational and cycle variance was guaranteed by altering the sensor and state conditions within each sequence and causing occasional anomalies within the process around 10% of the time. Table 2 in the appendix illustrates the performance of the root cause assignment using the ensemble model compared to its individual components (Incremental AutoEncoder, Principal Component Analysis and Mutual Information MI) as well as other machine learning techniques, namely One-Class Support Vector Machine (OC-SVM) [29], Isolation Forest [30], and k-Nearest Neighbors (kNN) [31]. This evaluation highlights a significant advantage in true positive detection for both the ensemble and the Incremental AutoEncoder over the other models. Due to the Incremental AE detecting way more false positives than the ensemble, the validation shows the benefit of the ensemble scoring over singular models. Nonetheless, there remains room for improvement in the ensemble's performance, as indicated by its F1-Score of 75.3%.

### 5. Conclusion

This research addresses productivity loss diagnostics in manufacturing systems through an ensemble machine learning framework. It offers a robust solution for pinpointing the root causes of inefficiencies, by leveraging cyclic multivariate time series data from available PLC data. The methods capacity for near-real-time analysis ensures a fast response in interventions against productivity issues, consequentially reducing manufacturing downtime and associated financial losses. By automating the root cause diagnosis process, the model not only accelerates the resolution of productivity issues but also reduces the reliance on extensive and expensive expert consultation, further driving efficiency and bottleneck resolving in the production process.

The proposed approach also opens up potential for further research and development. The individual parts of the ensemble can be revisited and optimized, perhaps identifying more capable modeling approaches. The Incremental AutoEncoder still leaves room for improvement, as the field of Deep Learning generates new insights with high frequency. Additionally, the hyperparameter configuration can be automated and adapted to the data to be analyzed, improving versatility and potentially even precision.

In order to further validate the framework and obtain feedback for future developments, an application in a wider range of manufacturing and automation systems should be carried out.

### Acknowledgements

This work was supported by the German Federal Government's 7th Energy Research Program "Innovationen für die Energiewende", under the project initiative "FlachMembranbefeuchterModul Industrialisierung (FLAMMI)". The authors express their gratitude for the financial support, resources and data provided.

# Appendix

## Data and Descriptive Analysis

Table 1: Cut-out from a manufacturing systems dataset. The dataset follows two key characteristics: a) Multivariate Binary Time Series: Sensor readings and PLC signals are time-stamped, allowing for temporal analysis. b) Cyclic: Each cycle contains a set of operations repeated over time, providing a structured framework for analysis

| Timestamp | Signal 1 | Signal 2 | Signal 3 | Signal 4 | Signal 5 | … | Cycle | State |
|---|---|---|---|---|---|---|---|---|
| 19:04:15.0684 | 0 | 1 | 1 | 1 | 1 | | 1 | 1 |
| 19:04:15.9605 | 0 | 1 | 1 | 1 | 1 | | 1 | 1 |
| 19:04:42.8403 | 1 | 0 | 0 | 0 | 0 | | 1 | 2 |
| 19:04:43.2353 | 1 | 0 | 0 | 0 | 0 | | 1 | 2 |
| 19:05:13.2559 | 1 | 0 | 0 | 1 | 1 | | 1 | 3 |
| 19:05:17.1166 | 1 | 0 | 0 | 0 | 1 | | 1 | 4 |
| 19:05:50.6370 | 0 | 1 | 0 | 0 | 1 | | 1 | 5 |
| 19:06:07.4969 | 1 | 0 | 0 | 1 | 0 | | 1 | 6 |
| 19:06:27.6087 | 0 | 0 | 0 | 0 | 1 | | 1 | 7 |
| 19:06:53.4058 | 0 | 0 | 1 | 1 | 1 | | 2 | 1 |
| 19:09:14.7522 | 0 | 1 | 1 | 1 | 0 | | 2 | 2 |

**Further visualization for the injection molding automation system in 4.2**

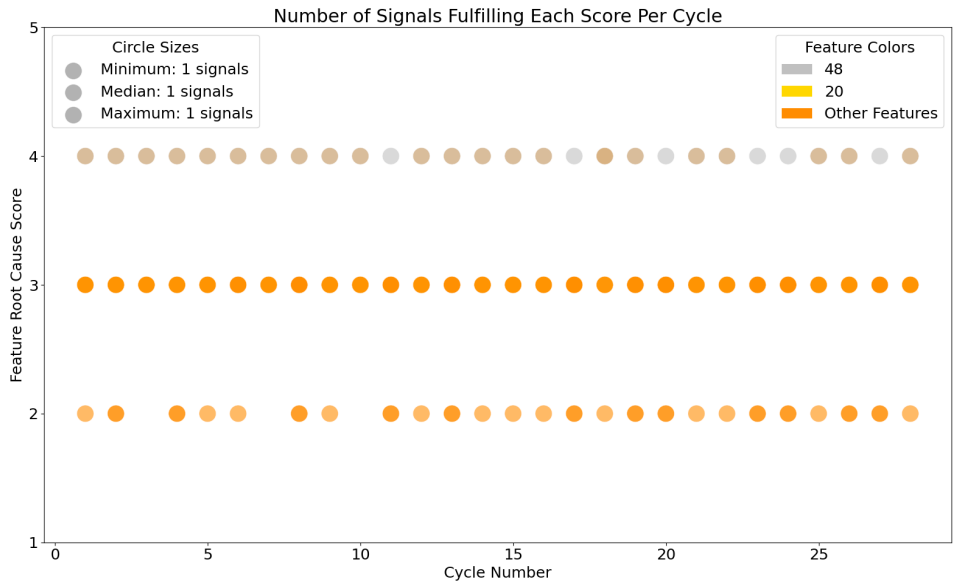

Figure 3: Root cause score assignment for non-zero scored features during the observed high loss cycles.

## Numerical evaluation of the proposed model on the 4.3 synthetic PLC data

Table 2: Numerical validation of the proposed ensemble model, its singular parts and three benchmark machine-learning models on the labeled synthetic PLC dataset.

| Model | F1-Score | Recall | Precision | True Positive | False Positive | False Negative |
|---|---|---|---|---|---|---|
| Ensemble | 0,753 | 0,695 | 0,822 | 430 | 93 | 189 |
| Incremental AE | 0,528 | 0,564 | 0,496 | 349 | 355 | 270 |
| PCA | 0,317 | 0,294 | 0,345 | 182 | 346 | 437 |
| MI | 0,194 | 0,179 | 0,210 | 111 | 417 | 508 |
| OC-SVM | 0,159 | 0,147 | 0,172 | 91 | 437 | 528 |
| Isolation Forest | 0,157 | 0,145 | 0,170 | 90 | 438 | 529 |
| KNN | 0,139 | 0,129 | 0,152 | 80 | 448 | 539 |

## Hyperparameters chosen for the validation

A few architectural details and hyperparameters have been necessary to conduct the above validation.

The most significant feature threshold $t$ mentioned in chapter 3 for selecting $I_{1a}^c$ and $I_{1b}^c$ has been chosen as the highest 3.75% of scores. For $I_2^c$ the highest 6.5% have been selected, for the PCA in $I_3^c$ the top 7.5% and for the XGBoost the top 10% of features.

The Autoencoder model implemented is comprised of three linear encoding layers with a non-linear ReLU activation function each. The latent dimension of neurons for the Encoder layer one is 32, for layer two 16 and for the final encoding layer 8. It includes L1 and L2 Regularization with coefficients of 0.01 to mitigate overfitting by penalizing large weights as well as a Dropout of 0.2 between layers. The training has been realized with two epochs per cycle batch and an additional five epochs per replay training.

The Decoder is reversed to the Encoder, being built from a linear layer mapping the encoded 8 Neurons back to 16 and finally 32, employing non-linear Activation, Regularization and Dropout mirroring the Encoder. The final output layer of the Decoder maps the data back to its original shape and is combined with a Sigmoid Activation Function. The ADAM algorithm was utilized as optimizer. The $\lambda$ weighing parameter of the EWC loss was chosen to be 0.4, the Fisher matrix was computed every 50 observed cycles.

For the XGBoost model training, logarithmic loss was selected.

The benchmark OC-SVM model was deployed with the 'rbf' kernel.

The Isolation Forest benchmark model was validated with an approximated anomaly contamination value of 0.1.

For the $k$ neighbors to be consider in the kNN algorithm, $k = 10$ was chosen.

## References

[1] Bauernhansl, T., 2020. Fabrikbetriebslehre 1, 1st ed. Springer, Berlin, Heidelberg.
[2] Slack, N., Brandon-Jones, A., Burgess, N., 2022. Operations management, 10th ed. Pearson, Harlow.
[3] Sai, B.K., Gram, J., Bauernhansl, T., 2023. Information-based Preprocessing of PLC Data for Automatic Behavior Modeling. Procedia CIRP 120, 565–571.
[4] Sai, B.K., Mayer, Y.T., Bauernhansl, T., 2021. Dynamic Data Acquisition and Preprocessing for Automatic Behavioral Modeling of Cyber-physical Systems. Procedia CIRP 104, 363–369.
[5] Nakajima, S., 1988. Introduction to TPM: Total productive maintenance, 1st ed. Productivity Press, Cambridge, Mass.

[6] Muchiri, P., Pintelon, L., 2008. Performance measurement using overall equipment effectiveness (OEE): literature review and practical application discussion. International Journal of Production Research 46 (13), 3517–3535.

[7] Tangen, S., 2003. An overview of frequently used performance measures. Work Study 52 (7), 347–354.

[8] Ungern-Sternberg, R., Leipoldt, C., Erlach, K., 2021. Work Center Performance Measurement Based On Multiple Time Series. Procedia CIRP 104, 276–282.

[9] Chen, X., Zheng, J., Zhao, C., Wu, M., 2024. Full Decoupling High-Order Dynamic Mode Decomposition for Advanced Static and Dynamic Synergetic Fault Detection and Isolation. IEEE Trans. Automat. Sci. Eng. 21 (1), 226–240.

[10] Perepu, S.K., Pinnamaraju, V.S., 2022. A novel unsupervised method for root cause analysis of anomalies using sparse optimization techniques, in: 2022 10th International Conference on Systems and Control (ICSC). 2022 10th International Conference on Systems and Control (ICSC), Marseille, France. 23.11.2022 - 25.11.2022. IEEE, pp. 416–422.

[11] Budhathoki, K., Minorics, L., Bloebaum, P., Janzing, D., 2022. Causal structure-based root cause analysis of outliers, in: Proceedings of the 39th International Conference on Machine Learning. PMLR, pp. 2357–2369.

[12] Assaad, C.K., Ez-zejjari, I., Zan, L., 2023. Root Cause Identification for Collective Anomalies in Time Series given an Acyclic Summary Causal Graph with Loops, in: Proceedings of The 26th International Conference on Artificial Intelligence and Statistics. PMLR, pp. 8395–8404.

[13] Wehner, C., Kertel, M., Wewerka, J., 2023. Interactive and Intelligent Root Cause Analysis in Manufacturing with Causal Bayesian Networks and Knowledge Graphs, in: 2023 IEEE 97th Vehicular Technology Conference (VTC2023-Spring). 2023 IEEE 97th Vehicular Technology Conference (VTC2023-Spring), Florence, Italy. 20.06.2023 - 23.06.2023. IEEE, pp. 1–7.

[14] Garrido-Labrador, J.L., Puente-Gabarri, D., Ramírez-Sanz, J.M., Ayala-Dulanto, D., Maudes, J., 2020. Using Ensembles for Accurate Modelling of Manufacturing Processes in an IoT Data-Acquisition Solution. Applied Sciences 10 (13), 4606.

[15] Jose, J.P., Ananthan, T., Prakash, N.K., 2022. Ensemble Learning Methods for Machine Fault Diagnosis, in: 2022 Third International Conference on Intelligent Computing Instrumentation and Control Technologies (ICICICT). 2022 Third International Conference on Intelligent Computing Instrumentation and Control Technologies (ICICICT), Kannur, India. 11.08.2022 - 12.08.2022. IEEE, pp. 1127–1134.

[16] Kaupp, L., Humm, B., Nazemi, K., Simons, S., 2022. Autoencoder-Ensemble-Based Unsupervised Selection of Production-Relevant Variables for Context-Aware Fault Diagnosis. Sensors (Basel, Switzerland) 22 (21).

[17] Zeng, Y., Chen, X., Jin, R., 2024. Ensemble Active Learning by Contextual Bandits for AI Incubation in Manufacturing. ACM Trans. Intell. Syst. Technol. 15 (1), 1–26.

[18] Farahani, M.A., McCormick, M.R., Gianinny, R., Hudacheck, F., Harik, R., Liu, Z., Wuest, T., 2023. Time-series pattern recognition in Smart Manufacturing Systems: A literature review and ontology. Journal of Manufacturing Systems 69, 208–241.

[19] Polikar, R., 2006. Ensemble based systems in decision making. IEEE Circuits Syst. Mag. 6 (3), 21–45.

[20] Goodfellow, I., Bengio, Y., Courville, A., 2016. Deep Learning. MIT Press, Cambridge, Massachusetts.

[21] McCloskey, M., Cohen, N.J., 1989. Catastrophic Interference in Connectionist Networks: The Sequential Learning Problem, in: , vol. 24. Elsevier, pp. 109–165.

[22] Kirkpatrick, J., Pascanu, R., Rabinowitz, N., Veness, J., Desjardins, G., Rusu, A.A., Milan, K., Quan, J., Ramalho, T., Grabska-Barwinska, A., Hassabis, D., Clopath, C., Kumaran, D., Hadsell, R., 2017. Overcoming catastrophic forgetting in neural networks. Proceedings of the National Academy of Sciences of the United States of America 114 (13), 3521–3526.

[23] Pascanu, R., Bengio, Y., 2014. Revisiting Natural Gradient for Deep Networks. International Conference on Learning Representations (ICLR).

[24] Parisi, G.I., Kemker, R., Part, J.L., Kanan, C., Wermter, S., 2019. Continual lifelong learning with neural networks: A review. Neural networks : the official journal of the International Neural Network Society 113, 54–71.

[25] Shannon, C.E., 1948. A Mathematical Theory of Communication. Bell System Technical Journal 27 (3), 379–423.

[26] Tipping, M.E., Bishop, C.M., 1999. Probabilistic Principal Component Analysis. Journal of the Royal Statistical Society Series B: Statistical Methodology 61 (3), 611–622.

[27] Friedman, J.H., 2001. Greedy function approximation: A gradient boosting machine. Ann. Statist. 29 (5).

[28] Chen, T., Guestrin, C., 2016. XGBoost, in: Proceedings of the 22nd ACM SIGKDD International Conference on Knowledge Discovery and Data Mining. KDD '16: The 22nd ACM SIGKDD International Conference on Knowledge Discovery and Data Mining, San Francisco California USA. 13 08 2016 17 08 2016. ACM, New York, NY, USA, pp. 785–794.

[29] Schölkopf, B., Platt, J.C., Shawe-Taylor, J., Smola, A.J., Williamson, R.C., 2001. Estimating the support of a high-dimensional distribution. Neural computation 13 (7), 1443–1471.

[30] Liu, F.T., Ting, K.M., Zhou, Z.-H., 2008. Isolation Forest, in: , Eighth IEEE International Conference on Data Mining, 2008. ICDM '08 ; Pisa, Italy, 15 - 19 Dec. 2008. IEEE, Piscataway, NJ, pp. 413–422.

[31] Zhang, Z., 2016. Introduction to machine learning: k-nearest neighbors. Annals of translational medicine 4 (11), 218.

**Biography**

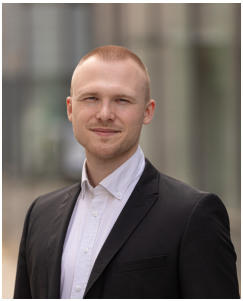

**Jonas Gram** (*1995), has been a Research Associate of the Department of Autonomous Production Optimization of the Fraunhofer Institute of Manufacturing Engineering and Automation IPA in Stuttgart since 2022. His research focus is on anomaly detection and root cause analysis in multivariate time series data, as well as data-driven modeling of complex and dynamic systems.

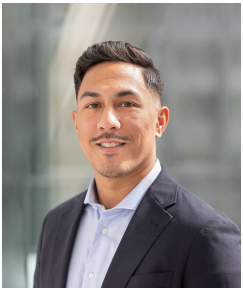

**Brandon Sai** (*1990) has been a Research Associate of the Department of Autonomous Production Optimization of the Fraunhofer Institute of Manufacturing Engineering and Automation IPA in Stuttgart and a doctoral student at the Institute of Industrial Manufacturing and Management at the University of Stuttgart since 2018. He is currently working on behavioral models of production systems.

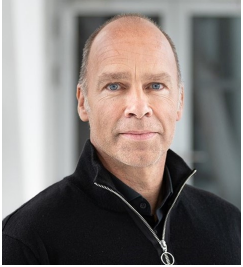

**Thomas Bauernhansl** (*1969) has been the Director of the Fraunhofer Institute for Manufacturing Engineering and Automation IPA in Stuttgart and the Institute of Industrial Manufacturing and Management at the University of Stuttgart since 2011. Both the digital and the biological transformation of value creation are core research fields in his institutes.